\def\year{2020}\relax
\documentclass[letterpaper]{article} % DO NOT CHANGE THIS
\usepackage{aaai20}  % DO NOT CHANGE THIS
\usepackage{times}  % DO NOT CHANGE THIS
\usepackage{helvet} % DO NOT CHANGE THIS
\usepackage{courier}  % DO NOT CHANGE THIS
\usepackage[hyphens]{url}  % DO NOT CHANGE THIS
\usepackage{graphicx} % DO NOT CHANGE THIS
\usepackage{graphicx}  % DO NOT CHANGE THIS
\nocopyright

\usepackage{algorithm}
\usepackage[noend]{algpseudocode}

\usepackage{xcolor}
\usepackage{comment}

\usepackage{arydshln}

\usepackage{tikz}
\usetikzlibrary{matrix,chains,decorations.pathreplacing,fit,calc}
\usetikzlibrary{patterns}
\usetikzlibrary{positioning,automata}
\usepackage{tikz-qtree}  
\usepackage{amsmath}
\usetikzlibrary{shapes,backgrounds}
\usetikzlibrary{positioning,shapes.geometric}
\usetikzlibrary{matrix,chains,decorations.pathreplacing,fit}
\usetikzlibrary{arrows}
\usetikzlibrary{intersections,positioning}
\usepackage{tikz-qtree-compat}

\definecolor{Gray}{gray}{0.9}
\newcommand{\blue}[1]{\textcolor{blue}{#1}}
\newcommand{\red}[1]{\textcolor{red}{#1}}

\newcommand{\redcolor}[1]{\textcolor{red}{#1}}
\newcommand{\upred}{\redcolor{\uparrow}}
\newcommand{\textuparrow}{{\uparrow}}
\newcommand{\downred}{\redcolor{\downarrow}}
\newcommand{\textdownarrow}{{\downarrow}}

\DeclareMathOperator*{\argmax}{argmax}

\newcommand{\ashish}[1]{\textcolor{blue}{[#1 \textsc{--ashish}]}}

\newcommand{\namecite}[1]{\citeauthor{#1}~\shortcite{#1}}
\newcommand{\citet}{\namecite}

\usepackage{listings}

\definecolor{codegreen}{rgb}{0,0.6,0}
\definecolor{codegray}{rgb}{0.5,0.5,0.5}
\definecolor{codepurple}{rgb}{0.58,0,0.82}
\definecolor{backcolour}{rgb}{0.95,0.95,0.92}

\newcommand\T{\rule{0pt}{2.6ex}}       % Top strut
\newcommand\B{\rule[-1.2ex]{0pt}{0pt}} % Bottom strut
\lstdefinestyle{mystyle}{
    backgroundcolor=\color{backcolour},   
    commentstyle=\color{codegreen},
    keywordstyle=\color{magenta},
    numberstyle=\tiny\color{codegray},
    stringstyle=\color{codepurple},
    basicstyle=\ttfamily\footnotesize,
    breakatwhitespace=false,         
    breaklines=true,                 
    captionpos=b,                    
    keepspaces=true,                 
    numbers=left,                    
    numbersep=5pt,                  
    showspaces=false,                
    showstringspaces=false,
    showtabs=false,                  
    tabsize=2
}

\title{Probing Natural Language Inference Models through Semantic Fragments}  % maybe 

\author{
\textbf{
  Kyle Richardson\textsuperscript{\rm $^\dag$}
  \and
  Hai Hu{\rm $^\ddag$}
  \and
  Lawrence S. Moss{\rm $^\ddag$}
  \and
  Ashish Sabharwal\textsuperscript{\rm $^\dag$}
} \\
 \\
\textsuperscript{\rm $\dag$}Allen Institute for AI, Seattle, WA, USA \\
\textsuperscript{\rm $\ddag$}Indiana University, Bloomington, IN, USA \\
\textsuperscript{\rm $\dag$}\{kyler,ashishs\}@allenai.org, \textsuperscript{\rm $\ddag$}\{huhai, lmoss\}@indiana.edu
}

\begin{document}

\maketitle

%Progress in natural language inference (NLI) has accelerated in recent years, due largely to the introduction of new large-scale datasets and advances in neural modeling. With these performance increases have come increased scrutiny of systematic annotation biases in benchmark datasets,

\begin{abstract}
Do state-of-the-art models for language understanding already have, or can
they easily learn, abilities such as boolean coordination, quantification, conditionals, comparatives, and monotonicity reasoning (i.e., reasoning about word substitutions in sentential contexts)? While such phenomena are involved in natural language inference (NLI) and go beyond basic linguistic understanding, it is unclear the extent to which they are captured in existing NLI benchmarks and effectively learned by  models. To investigate this, we propose the use of \emph{semantic fragments}---systematically generated datasets that each target a different semantic phenomenon---for probing, and efficiently improving, such capabilities of linguistic models. This approach to creating challenge datasets allows direct control over the semantic diversity and complexity of the targeted linguistic phenomena, and results in a more precise characterization of a model's linguistic behavior. 
% kyle : i took this out, sounds a bit risky to me
%Indeed, extreme instances in our fragments are so complex that even a typical human would struggle with them. Yet, they are, by design, linguistically and logically sound.
% , resulting in a precise characterization of a model's linguistic behavior.
Our experiments, using a library of 8 such semantic
% fragments involving logic and monotonicity reasoning,
fragments, reveal two remarkable findings: (a) State-of-the-art models, including BERT, that are pre-trained on existing NLI benchmark datasets perform poorly on these new fragments, even though the phenomena probed here are central to the NLI task; (b) On the other hand, with only a few minutes of additional fine-tuning---with a carefully selected learning rate and a novel variation of ``inoculation''---a BERT-based model can master all of these logic and monotonicity fragments while retaining its performance on established NLI benchmarks. 

\end{abstract}

%% introduction 
\section{Introduction}
\label{sec:intro}

%The goal of natural language inference (NLI) is to develop models that can detect inferential relations between pairs of text. For example, in Figure~\ref{fig:first_example} 
%\peter{I still would recommend dropping this first paragraph completely, it's unnecessary background, while you should be making your main case for the paper. You could move it later if necessary. For the first mention of montonicity reading in the subsequent paragraph, I think just a footnote or parenthetical would suffice for explanation, e.g., "...monotonicity inference (i.e., truth-preserving substitution of a more general or specific term)"}

\begin{comment}
\ashish{added an opening para that may align better with the new abstract}
\end{comment} 

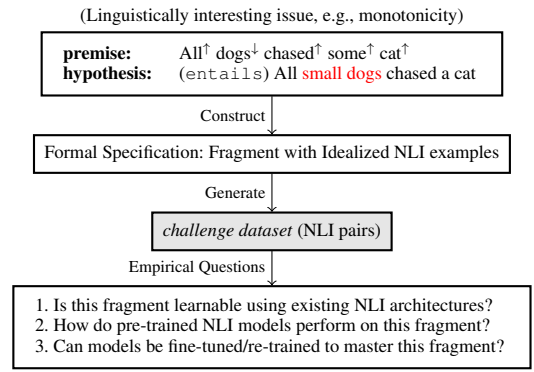
\begin{figure}[t]
    \centering
 \begin{tikzpicture}[scale=0.65, every node/.style={scale=0.7}]
\tikzstyle{source} = [draw,thick,inner sep=.2cm,fill=Gray];
\tikzstyle{source2} = [draw,thick,inner sep=.2cm];
\tikzstyle{process} = [draw,thick,circle]; 
\tikzstyle{to} = [-,shorten >=1pt,semithick,font=\sffamily\footnotesize];
\matrix (compiler) 
[matrix of nodes,
  ampersand replacement=\&,
  row sep=1cm,
  column sep=-.9cm,
  nodes={align=left},
  ]
{
 \node[source2,label={\normalsize (Linguistically interesting issue, e.g., monotonicity)}] (sa) {
 		\begin{tabular}{l l} 
		\textbf{premise:} & All$^{\uparrow}$  dogs$^{\downarrow}$ chased$^{\uparrow}$ some$^{\uparrow}$ cat$^{\uparrow}$ \\
		\textbf{hypothesis: } & (\texttt{entails}) All \textcolor{red}{small dogs} chased a cat 	
 		\end{tabular}
 };  \\[-.5cm]
 %\& \node[source](bert) {SOTA NLI Model}; \\[-.6cm]
  \node[source2] (rules) {Formal Specification: Fragment with Idealized NLI examples}; \\[-.5cm]
  \node[source] (programs){\emph{challenge dataset} (NLI pairs)}; \& \node[](dummy){}; \\[-.55cm]
  \node[source2](result){
  	\begin{tabular}{l} 
		1. Is this fragment learnable using existing NLI architectures? \\
		2. How do pre-trained NLI models perform on this fragment? \\
		3. Can models be fine-tuned/re-trained to master this fragment? 
	\end{tabular}
  }; \\[-1.3cm]
}; 
\draw[->] (sa) -- node[left] {\footnotesize Construct} (rules); 
\draw[->] (rules) -- node[left] {\footnotesize Generate} (programs); 
\draw[->] (programs) -- node[left] {\footnotesize Empirical Questions} (result);
%\draw[->,thick] (bert) -- (programs.east);
%\draw[->,dashed,thick] (programs) to [out=30,in=40, looseness=2] node[right ]{\textcolor{red}{inoculate}} (bert);
\end{tikzpicture} 

\caption{An illustration of our proposed method for studying NLI model behavior through \emph{semantic fragments}.}
    \label{fig:first_example}
\end{figure}

Natural language inference (NLI) is the task of detecting inferential relationships between natural language descriptions. For example, given the pair of sentences \emph{All dogs chased some cat} and \emph{All small dogs chased a cat} shown in Figure~\ref{fig:first_example}, the goal for an NLI model is to determine that the second sentence, known as the \textbf{hypothesis} sentence, follows from the meaning of the first sentence (the \textbf{premise} sentence). Such a task is known to involve a wide range of reasoning and knowledge phenomena, including knowledge that goes beyond basic linguistic understanding (e.g., elementary logic). As one example of such knowledge, the \emph{inference} in Figure~\ref{fig:first_example} involves monotonicity reasoning (i.e., reasoning about word substitutions in context); here the position of \emph{dogs} in the premise occurs in a \emph{downward monotone} context (marked as $\downarrow$), meaning that it can be \emph{specialized} (i.e., substituted with a more specific concept such as \emph{small dogs}) to generate an entailment relation. In contrast, substituting \emph{dogs} for a more generic concept, such as \emph{animal}, has the effect of generating a \texttt{NEUTRAL} inference.

In an empirical setting, it is desirable to be able to measure the extent to which a given model captures such types of knowledge. We propose to do this using a suite of controlled dataset probes that we call \emph{semantic fragments}.

\begin{comment}
\ashish{see comment where this figure is referenced; doesn't quite work/help in the current form. according to the description, the topmost box should have a generic rule template (expecting a line from Fig~\ref{fig:logic_templates}) and the middle box should have specific examples (similar to dogs and cats). The "generate" item should also have some access to data / entities, etc., to ground the rules. Let's figure out a better figure that will help readers more!}
\end{comment}

% , among others \cite{khot2018scitail}

While NLI has long been studied in linguistics and logic and has focused on specific types of logical phenomena such as monotonicity inference, attention to these topics has come only recently to empirical NLI. Progress in empirical NLI has accelerated due to the introduction of new large-scale  NLI datasets, such as the Stanford Natural Language Inference (SNLI) dataset \cite{snli} and MultiNLI (MNLI) \cite{mnli}, coupled with new advances in neural modeling and model pre-training \cite{conneau2017supervised,devlin2018bert}. With these performance increases has come increased scrutiny of systematic annotation biases in existing datasets \cite{poliak2018hypothesis,gururangan2018annotation}, as well as attempts to build new \emph{challenge datasets} that focus on particular linguistic phenomena \cite{glockner2018breaking,naik2018stress,poliak2018collecting}. The latter aim to more definitively answer questions such as: are models able to effectively learn and extrapolate complex knowledge and reasoning abilities when trained on benchmark tasks?

To date, studies using challenge datasets have largely been limited by the simple types of inferences that they included (e.g., lexical and negation inferences). They fail to cover more complex reasoning phenomena related to logic, and primarily use adversarially generated corpus data, which sometimes makes it difficult to identify exactly the particular semantic phenomena being tested for. There is also a focus on datasets that are easily able to be constructed and/or verified using crowd-sourcing techniques. Adequately evaluating a model's \emph{competence} on a given reasoning phenomena, however, often requires datasets that are hard even for humans, but that are nonetheless based on sound formal principles (e.g., reasoning about monotonicity where, in contrast to the simple example in Figure~\ref{fig:first_example}, several nested downward monotone contexts are involved to test the model's capacity for compositionality, cf.~\citet{lake2017generalization}).

%**There is also a focus on entailment problems that are easy for humans; we instead focus on some of the reasoning problems that are difficult for humans, but that are nonetheless logically justified** \textcolor{red}{this needs modification}\footnote{LM:
%here is a suggestion for the point that needs modification:
%To probe whether a neural learner is truly expert in some task, we should examine it with 
%both easy and hard questions.   We aim to learn whether the learner can extrapolate
%from humanly-easy examples to ones which exceed what people typically do. 
%This is what motivates our use of \emph{semantic fragments}.
%}%

%the  fragment lack complexity. What's missing is a dquadefinitive proof, such as montonicity reasoning (i.e., the type of reasoning discussed above). 

\begin{figure*}
\setlength{\doublerulesep}{\arrayrulewidth}
\centering
\scalebox{0.83}{
{\footnotesize
\begin{tabular}{| l l l c c c |}
\multicolumn{1}{c}{\B \textbf{Fragments}} & \multicolumn{1}{c}{ \textbf{Example }(\texttt{premise},\texttt{label},\texttt{hypothesis})} & \multicolumn{1}{c}{\textbf{Genre}} & \multicolumn{1}{c}{\textbf{Vocab. Size}} & \multicolumn{1}{c}{\textbf{\# Pairs}} & \multicolumn{1}{c}{\textbf{Avg. Sen. Len.}} \\ \hline\hline
\texttt{Negation} & \begin{tabular}{l} \T \emph{Laurie has only visited Nephi, Marion has only visited Calistoga.} \\
\texttt{CONTRADICTION}\emph{ Laurie didn't visit Calistoga}
\end{tabular} & Countries/Travel & 3,581 & 5,000 & 20.8 \\ \hline
\texttt{Boolean} & \begin{tabular}{l} \T \emph{Travis, Arthur, Henry and Dan have only visited Georgia}   \\
\texttt{ENTAILMENT }\emph{Dan didn't visit Rwanda}
\end{tabular} & Countries/Travel & 4,172 & 5,000 & 10.9 \\ \hline
\texttt{Quantifier} & \begin{tabular}{l} \T \emph{Everyone has visited every place} \\
\texttt{NEUTRAL}\emph{ Virgil didn't visit Barry}
\end{tabular} & Countries/Travel & 3,414 & 5,000 & 9.6 \\ \hline
\texttt{Counting} & \begin{tabular}{l} \T \emph{Nellie has visited Carrie, Billie, John, Mike, Thomas, Mark, .., and Arthur.} \\
\texttt{ENTAILMENT}\emph{ Nellie has visited more than 10 people.}
\end{tabular} & Countries/Travel & 3,879 & 5,000 & 14.0 \\ \hline
\texttt{Conditionals} & \begin{tabular}{l} \T \emph{Francisco has visited Potsdam and if Francisco has visited Potsdam} \\
\emph{then Tyrone has visited Pampa} \texttt{ENTAILMENT}\emph{ Tyrone has visited Pampa.}
\end{tabular} & Countries/Travel & 4,123 & 5,000 & 15.6 \\ \hline
\texttt{Comparatives} & \begin{tabular}{l} \T \emph{John is taller than Gordon and Erik..., and Mitchell is as tall as John} \\
\texttt{NEUTRAL}\emph{ Erik is taller than Gordon.}
\end{tabular} & People/Height & 1,315 & 5,000 & 19.9 \\ \hline\hline
\texttt{Monotonicity} & \begin{tabular}{l} \T \emph{All black mammals saw exactly 5 stallions who danced }\texttt{ENTAILMENT} \\
\emph{A brown or black poodle saw exactly 5 stallions who danced}
\end{tabular} & Animals & 119 & 10,000 & 9.38 \\  \hline\hline % \hline\hline
\texttt{SNLI+MNLI} & \begin{tabular}{l} \T \emph{During calf roping a cowboy calls off his horse. } \\
\emph{\texttt{CONTRADICTION }A man ropes a calf successfully.}
\end{tabular} & Mixed & 101,110 & 942,069 & 12.3 \\ \hline
\end{tabular}}}

%\vspace{.3cm}
\caption{Information about the semantic fragments considered in this paper, where the top four fragments test basic logic (\texttt{Logic Fragments}) and the last fragment covers monotonicity reasoning (\texttt{Mono.~Fragment}).}
\label{fig:frags}
\end{figure*}

% , of the sort used in linguistic semantics,
%, of the sort used in linguistic semantics,

In contrast to existing work on challenge datasets, we propose using \emph{semantic fragments}---synthetically generated challenge datasets, of the sort used in linguistics, to study NLI model behavior. Semantic fragments provide the ability to systematically control the semantic complexity of each new challenge dataset by bringing to bear the expert knowledge excapsulated in formal theories of reasoning, making it possible to more precisely identify model performance and competence on a given linguistic phenomenon. While our idea of using fragments is broadly applicable to any linguistic or reasoning phenomena, we look at eight types of fragments that cover several fundamental aspects of reasoning in NLI, namely, monotonicity reasoning using two newly constructed challenge datasets as well as  six other fragments that probe into rudimentary logic using new versions of the data from \citet{salvatore2019using}.  
%., all of which are fundamental to more complex forms of reasoning. 

\begin{comment}
\ashish{The statements/claims above are rather abstract. This would be a good place to ground them by (a) giving an example of what's in these fragments---something that would seem likely not covered by existing `challenge datasets' cited above; and (b) mentioning that these fragments can get less `natural' and quite complex; some extreme cases are difficult even for a typical human, e.g., XYZ [add a convincing example]; yet, by design (why/how exactly??), these extreme cases are still linguistically accurate and logically sound; make the case (in one sentence) that even these extreme cases probe NLI models in ways the community would want (i.e., the work is useful even though the sentences may not seem natural).}
\end{comment}

%\footnote{All data and code discussed in this paper will be publicly released upon publication.}.
\begin{comment}
\ashish{the topmost box in Fig~\ref{fig:first_example} doesn't quite with this description, so referring to the figure gets confusing}
\end{comment}

As illustrated in Figure~\ref{fig:first_example}, our proposed method works in the following way: starting with a particular linguistic fragment of interest, we create a formal specification (or a formal rule system with certain guarantees of correctness) of that fragment, with which we then automatically generate a new \emph{idealized} challenge dataset, and ask the following three empirical questions. 1) Is this particular fragment learnable from scratch using existing NLI architectures (if so, are the resulting models useful)?  2) How well do large state-of-the-art pre-trained NLI models (i.e., models trained on all known NLI data such as SNLI/MNLI) do on this task? 3) Can existing models be \emph{quickly} re-trained or re-purposed to be robust on these fragments (if so, does mastering a given linguistic fragment affect performance on the original task)?

We emphasize the \emph{quickly} part in the last question; given the multitude of possible fragments and linguistic phenomena that can be formulated and that we expect a wide-coverage NLI model to cover, we believe that models should be able to efficiently learn and adapt to new phenomena as they are encountered without having to learn entirely from scratch. In this paper we look specifically at the question: are there particular linguistic fragments (relative to other fragments) that are hard for these pre-trained models to adapt to or that confuse the model on its original task? 

%We believe that this last question has largely been missing in recent NLI studies; given the multitude of possible fragments and linguistic phenomena we would want a given NLI model to master, we expect that a given model should  

%Our investigation involves answering this question using the state-of-the-art BERT architecture and fine-tuning approach from \citet{devlin2018bert}, as well as two additional task-specific neural NLI architectures \cite{chen2017enhanced,parikh2016decomposable}.  

%, given its recent state-of-the-art performance on NLI benchmarks. 
On these eight fragments,
% and eight new challenge tasks,
we find that while existing NLI architectures can effectively learn these particular linguistic pheneomena, pre-trained NLI models do not perform well. This, as in other studies \cite{glockner2018breaking}, reveals weaknesses in the ability of these models to generalize. While most studies into linguistic probing end the story there, we take the additional step to see if attempts to continue the learning and re-fine-tune these models on fragments (using a novel and cheap \emph{inoculation} \cite{liu2019inoculation} strategy) can improve performance. Interestingly, we show that this yields mixed results depending on the particular linguistic phenomena and model being considered. For some fragments (e.g., comparatives), re-training some models comes at the cost of degrading performance on the original tasks, whereas for other phenomena (e.g., monotonicity) the learning is more stable, even across different models. These findings, and our technique of obtaining them, make it possible to identify the degree to which a given linguistic phenomenon \emph{stresses} a benchmark NLI model, and suggest a new methodology for quickly making models more robust.

%% expanded related work section 
\section{Related Work} 

The use of semantic fragments has a long tradition in logical semantics, starting with the seminal work of \citet{montague1973proper}, as well as earlier work on NLI \cite{fracas}.  We follow \citet{pratt2004fragments} in defining a \emph{semantic fragment} more precisely as a subset of a language \emph{equipped with semantics which translate sentences in a formal system such as first-order logic}. In contrast to work on empirical NLI, such linguistic work often emphasizes the complex cases of each phenomena in order measure \emph{competence} (see \citet{chomsky2014aspects} for a discussion about \emph{competence} vs. \emph{performance}). For our fragments that test basic logic, the target formal system includes basic boolean algebra, quantification, set comparisons and counting (see Figure~\ref{fig:frags}), and builds on the datasets from \citet{salvatore2019using}. For our second set of fragments that focus on monotonicity reasoning, the target formal system is based on the \emph{monotonicity calculus} of \citet{vanBenthemEssays86} (see review by \citet{IcardMoss2014}). To construct these datasets, we build on recent work on automatic polarity projection  \cite{hu2018polarity,HuChenMoss,hu2019monalog}. 

Our work follows other attempts to learn neural models from fragments and small subsets of language, which includes work on  syntactic probing \cite{mccoy2019right,goldberg2019assessing}, probing basic reasoning \cite{weston2015towards,geiger2018stress,geiger2019posing} and probing other tasks \cite{lake2017generalization,chrupala2019correlating,warstadt2019investigating}. \citet{geiger2018stress} is the closest work to ours.  However, they intentionally focus on artificial fragments that deviate from ordinary language, whereas our fragments (despite being automatically constructed and sometimes a bit pedantic) aim to test naturalistic subsets of English. In a similar spirit, there have been other attempts to collect datasets that target different types of inference phenomena \cite{white2017inference,poliak2018collecting}, which have been limited in linguistic complexity. Other attempts to study complex phenomena such as monotonicity reasoning in NLI models has been limited to training data augmentation \cite{yanaka2019help}, whereas we create several new challenge test sets to directly evaluate NLI performance on each phenomenon (see \citet{yanaka2019can} for closely related work that appeared concurrently with our work).  

%\cite{wang2018glue}

Unlike existing work on building NLI challenge datasets \cite{glockner2018breaking,naik2018stress}, we focus on the trade-off between mastering a particular linguistic fragment or phenomena independent of other tasks and data (i.e., Question 1 from Figure~\ref{fig:first_example}), while also maintaining performance on other NLI benchmark tasks (i.e., related to Question 3 in Figure~\ref{fig:first_example}).  To study this, we introduce a novel variation of the \emph{inoculation through fine-tuning} methodology of \citet{liu2019inoculation}, which emphasizes maximizing the model's \emph{aggregate} score over multiple tasks (as opposed to only on challenge tasks). Since our new challenge datasets focus narrowly on particular linguistic phenomena, we take this in the direction of seeing more precisely the extent to which a particular linguistic fragment stresses an existing NLI model. In addition to the task-specific NLI models looked at in  \citet{liu2019inoculation}, we  inoculate with the state-of-the-art pre-trained BERT model, using the fine-tuning approach of \citet{devlin2018bert}, which itself is based on the  transformer architecture of \citet{vaswani2017attention}.

\section{Some Semantic Fragments}

As shown in Figure~\ref{fig:first_example}, given a particular semantic fragment or linguistic phenomenon that we want to study, our starting point is a formal specification of that fragment (e.g., in the form of a set of templates/formal grammar that encapsulate expert knowledge of that phenomenon), which we can  sample
in order to obtain a new challenge set.  In this section, we describe the construction of the particular fragments we investigate in this paper, which are illustrated in Figure~\ref{fig:frags}. While these particular fragments seem to capture many of the core phenomena involved in NLI, we emphasize that any arbitrary linguistic fragment of interest could be constructed and subjected to the sets of experiments we describe in the next section. 

%We emphasize again that while we focus on a particular set of linguistic phenomena, our general approach is amenable to whichever linguistic fragment the experimenter wants to probe. 

%In this section, we describe the details of the different fragments we use in our studies, which are illustrated in Figure~\ref{fig:frags}. In each case, the starting point 

%Formulation here of semantic fragments; brief discussion of connection with linguistic background fragments in the tradition of \cite{montague1973proper}. 

\paragraph{The Logic Fragments} The first set of fragments probe into problems involving rudimentary logical reasoning. Using a fixed vocabulary of people and place names, individual fragments cover \emph{boolean coordination} (\texttt{boolean} reasoning about conjunction \texttt{and}), simple \texttt{negation}, quantification and quantifier scope (\texttt{quantifier}), \texttt{comparative} relations, set \texttt{counting}, and \texttt{conditional} phenomena all related to a small set of traveling and height relations.

These fragments (with the exception of the conditional fragment, which was built specially for this study) were first built using the set of verb-argument templates first described in \citet{salvatore2019using}.  Since their original rules were meant for 2-way NLI classification (i.e., \texttt{ENTAILMENT} and \texttt{CONTRADICTION}), we re-purposed their rule sets to handle 3-way classification, and added other inference rules, which resulted in some of the simplified templates shown in Figure~\ref{fig:logic_templates}. For each fragment, we uniformly generated 3,000 training examples and reserved 1,000 examples for testing. As in \citet{salvatore2019using}, the people and place names for testing are drawn from an entirely disjoint set from training.
We also reserve  1,000 for development. While we were capable of generating more data, we follow \citet{weston2015towards} in limiting the size of our training sets to 3,000 since our goal is to learn from as little data as possible, and found 3,000 training examples to be sufficient for most fragments and models. 

\begin{figure*}
\centering
\setlength{\doublerulesep}{\arrayrulewidth}
\scalebox{0.86}{
{\scriptsize
\begin{tabular}{| c l | }
\hline
     \textbf{Logic Fragment} & \multicolumn{1}{c|}{\textbf{Rule Template:} [ \textbf{premise} ], \{ \textbf{hypothesis}$_{1}$, ... \} $\Rightarrow \textbf{label}$; Labeled Examples (simplified)}  \\ \hline \hline
     %\\ \hline 
     \texttt{Negation} & 
            \begin{tabular}{l l l} 
                \texttt{[only-did-p($x$)], $\neg$p($x$) }  & $\Rightarrow$ \texttt{CONTRADICTION} &  \colorbox{Gray}{Dave$_{x}$} has only \colorbox{Gray}{visited Israel$_{p}$}, \colorbox{Gray}{Dave$_{x}$} \underline{didn't}$_{\neg}$ \colorbox{Gray}{visit Israel$_{p}$} \\
                \texttt{[only-did-p($x$)], $\neg$p'($x$) }  & $\Rightarrow$ \texttt{ENTAILMENT} & \colorbox{Gray}{Dave$_{x}$} has only \colorbox{Gray}{visited Israel$_{p}$}, \colorbox{Gray}{Dave$_{x}$} \underline{didn't}$\neg$ \colorbox{Gray}{visit Russia$_{p'}$} \\
                \texttt{[only-did-p($x$)], $\neg$p($x'$) }  & $\Rightarrow$ \texttt{NEUTRAL} &  \colorbox{Gray}{Dave$_{x}$} has only \colorbox{Gray}{visited Israel$_{p}$}, \colorbox{Gray}{Bill$_{x}$} \underline{didn't}$_{\neg}$ \colorbox{Gray}{visit Israel$_{p}$} \\
            \end{tabular} 
            \\ \hline 
    \texttt{Boolean} & 
            \begin{tabular}{l l l} 
                %\texttt{(p $\to$ q) $\land$ p, q } & $\Rightarrow$ \texttt{ENTAILMENT} \\
                \texttt{[p$(x_{1}) \land ... \land $p$(x_{n})$], \texttt{$\neg$p$(x_{j})$}}  & $\Rightarrow$ \texttt{CONTRADICTION} & \colorbox{Gray}{Dustin$_{x_{1}}$}, \colorbox{Gray}{Milton$_{x_{2}}$}, ... have \underline{only} \colorbox{Gray}{visited Equador$_{p}$}; \colorbox{Gray}{Dustin$x_{1}$} didn't$_{\neg}$ \colorbox{Gray}{visit Equador$_{p}$}  \\
                %\texttt{[p$_{1}(x) \land ... \land $p$_{n}(x)$], \texttt{$\neg$p$_{j}(x_{j})$}}  & $\Rightarrow$ \texttt{CONTRADICTION} \quad \colorbox{Gray}{Dustin$_{x}$} only \underline{visited$_{p}$} \colorbox{Gray}{Portugal$_{1}$}, \colorbox{Gray}{Spain$_{2}$},..; \colorbox{Gray}{Dustin$_{x}$} didn't$_{\neg}$ visit$_{p}$ \colorbox{Gray}{Spain$_{2}$}  \\
                \texttt{[p$_{1}(x_{1}) \land ... \land $p$_{n}(x_{n})$], \texttt{$\neg$p$_{j}(x')$}}  & $\Rightarrow$ \texttt{NEUTRAL} & \colorbox{Gray}{Dustin$_{x}$} only \underline{visited$_{p}$} \colorbox{Gray}{Portugal$_{1}$} and \colorbox{Gray}{Spain$_{2}$}; \colorbox{Gray}{James$_{x'}$} didn't$_{\neg}$ visit$_{p}$ \colorbox{Gray}{Spain$_{'}$} \\
                \texttt{[p$_{1}(x) \land ... \land $p$_{n}(x)$], \texttt{$\neg$p${}'(x)$}}  & $\Rightarrow$ \texttt{ENTAILMENT} & \colorbox{Gray}{Dustin$_{x}$} only \underline{visited$_{p}$} \colorbox{Gray}{Portugal$_{1}$} and \colorbox{Gray}{Spain$_{2}$}; \colorbox{Gray}{Dustin$_{x}$} didn't$_{\neg}$ visit$_{p}$ \colorbox{Gray}{Germany$_{'}$} \\
                %\texttt{(p $\to$ q) $\land$ $\neg$p, \{q,$\neg$q\}} & $\Rightarrow \texttt{NEUTRAL}
            \end{tabular} \\ \hline
     \texttt{Conditional} & 
            \begin{tabular}{l l l} 
                \texttt{[(p $\to$ q) $\land$ p], q } & $\Rightarrow$ \texttt{ENTAILMENT} & \colorbox{Gray}{Dave visited Israel$_{p}$} and if \colorbox{Gray}{Dave visited Israel$_{p}$} \underline{then$_{\to}$} \colorbox{Gray}{Bill visited Russia$_{q}$}; \colorbox{Gray}{Bill visited Russia$_{q}$}. \\
                \texttt{[(p $\to$ q) $\land$ p], $\neg$q } & $\Rightarrow$ \texttt{CONTRADICTION} & \colorbox{Gray}{Dave visited Israel$_{p}$} and if \colorbox{Gray}{Dave visited Israel$_{p}$} \underline{then$_{\to}$} \colorbox{Gray}{Bill visited Russia$_{q}$}; \colorbox{Gray}{Bill didn't visit Russia$_{p}$}. \\
                \texttt{[(p $\to$ q) $\land$ $\neg$p], \{q,$\neg$q\}} & $\Rightarrow$ \texttt{NEUTRAL}  & \colorbox{Gray}{Dave didn't visit Israel$_{p}$}, and if \colorbox{Gray}{Dave visited Israel$_{p}$} \underline{then$_{\to}$} \colorbox{Gray}{Bill visited Russia$_{q}$}; \colorbox{Gray}{Bill visited Russia$_{p}$}.
            \end{tabular} \\ \hline 
     \texttt{Quantifier} & 
            \begin{tabular}{l l l} 
            %\texttt{$[\forall x.$ p($x$)], \{$\exists x.\neg$p($x$), $\iota x.\neg$p($x$)\} }  & $\Rightarrow$ \texttt{CONTRADICTION} \\
            %\texttt{$[\forall x.$ p($x$)], \{$\exists x.$ p($x$), $\iota x.$ p($x$)\}}  & $\Rightarrow$ \texttt{ENTAILMENT} \\
            \texttt{$[\forall x. \forall y.$ p($x,y$)], $\exists x. \iota y.$ $\neg$p($x,y$) }  & $\Rightarrow$ \texttt{CONTRADICTION} & Everyone$_{\forall x}$ \colorbox{Gray}{visited$_{p}$} every$_{\forall}$ \colorbox{Gray}{country$_{y}$}; Someone$_{\exists x}$ \underline{didn't}$_{\neg}$ \colorbox{Gray}{visit$_{p}$} \colorbox{Gray}{Jordan$_{\iota y}$} \\
            \texttt{$[\exists x. \forall y.$ p($x,y$)], $\iota x. \exists y.$ \{$\neg$p($x,y$),p($x,y$)\} }  & $\Rightarrow$ \texttt{NEUTRAL} & Someone$_{\exists x}$ \colorbox{Gray}{visited$_{p}$} every$_{\forall}$ \colorbox{Gray}{person$_{y}$}; Tim$_{\iota x}$ \underline{didn't}$_{\neg}$ \colorbox{Gray}{visit$_{p}$} \colorbox{Gray}{someone$_{\exists y}$} \\
            \texttt{$[\exists x. \forall y.$ p($x,y$)], $\exists x. \iota y.$ p($x,y$) }  & $\Rightarrow$ \texttt{ENTAILMENT} & Someone$_{\exists x}$ \colorbox{Gray}{visited$_{p}$} every$_{\forall}$ \colorbox{Gray}{person$_{y}$}; A person$_{\exists x}$  \colorbox{Gray}{visited$_{p}$} \colorbox{Gray}{Mark$_{\iota y}$} \\ 
            %\texttt{(p $\to$ q) $\land$ $\neg$p, \{q,$\neg$q\}} & $\Rightarrow \texttt{NEUTRAL}
            \end{tabular} \\ \hline 
     %\hline \hline 
     %\texttt{Counting} & \begin{tabular}{l l} 
     %\texttt{[only(p$_{1}(x)$ $..\land..$ p$_{n}(x)$)], \{p$_{1}(x) \land..$p$_{n+1}(x)$, $n < n+j$ \}}  & $\Rightarrow$ %\texttt{CONTRADICTION} \\
     %\texttt{[p$_{1}(x)$ $..\land..$ p$_{n}(x)$], \{$n < n+j$, $n > n-j$ \}} & $\Rightarrow$ \texttt{ENTAILMENT} \\
     %\texttt{[p($x$) $\land\text{ }|\texttt{p}| = n$], $\iota x. $p($x'$) }  & $\Rightarrow$ \texttt{NEUTRAL} \\
     %\end{tabular}
     %\\ \hline 
     %\texttt{Comparative} & \begin{tabular}{l l} 
     %\texttt{[p$_{1}$ > p$_{2}$ ... > p$_{n},$, p$_{n}=$p$_{n+1}$], p$_{j}$ > p$_{j+1}$}  & $\Rightarrow$ \texttt{CONTRADICTION} %\\
     %\texttt{[p$_{1}$ > p$_{2}$ ... > p$_{n}$], p$_{j+1}$ > p$_{j}$}  & $\Rightarrow$ \texttt{ENTAILMENT} \\
     %\texttt{[p$_{1}$ > p$_{2}$; p$_{1}$ > p$_{3}$]], \{ p$_{2}$ > p$_{3}$, p$_{3}$ = p$_{3}$\}}  & $\Rightarrow$ \texttt{NEUTRAL} %\\
     %\end{tabular}
     %\\ \hline 
\end{tabular}}}
\caption{A simplified description of some of the templates used for 4 of the logic fragments (stemming from \citet{salvatore2019using}) expressed in a quasi-logical notation with predicates \texttt{p,q,only-did-p} and quantifiers $\exists\, (\text{there exists}),\forall\, (\text{for all}),\iota\, (\text{there exists a unique})$ and boolean connectives ($\land$ (and), $\to$ (if-then), $\neg$ (not)).} %and comparison operators ($>$ (greater-than) and $<$ (less-than)).}
\label{fig:logic_templates}
\end{figure*}

As detailed in Figure~\ref{fig:frags}, these new fragments vary in complexity, with the \texttt{negation} fragment (which is limited to verbal negation) being the least complex in terms of linguistic phenomena. We also note that all other fragments include basic negation and boolean operators, which we found to help preserve the naturalness of the examples in each fragment. As shown in last column of Figure~\ref{fig:frags}, some of our fragments (notably, \texttt{negation} and \texttt{comparatives}) have, on average, sentence lengths that exceed that of benchmark datasets. This is largely due to the productive nature of some of our rules. For example, the \texttt{comparatives} rule set allows us to create arbitrarily long sentences by generating long lists of people that are being compared (e.g., In \emph{John is taller than ..}, we can list up to 15 people in the subsequent list of people). %that John is \emph{taller than}).

Whenever creating synthetic data, it is important to ensure that one is not introducing into the rule sets particular annotation artifacts \cite{gururangan2018annotation} that make the resulting challenge datasets trivially learnable. As shown in the top part of Table~\ref{table:frag2_results}, which we discuss later, we found that several strong baselines failed to solve our fragments, showing that the fragments, despite their simplicity and constrained nature, are indeed not trivial to solve.  

%, which we discuss
%in Sections~\ref{sec:exp}-\ref{sec:findings}
%\textcolor{red}{fill this in}

%as well as with the monotonicity fragments discussed next. 
%We note that 

\begin{figure}[t]
	\centering
	\scalebox{0.8}{
		\begin{tikzpicture}
		\tikzset{align=center,level distance=65,sibling distance=30pt}
		\tikzset{edge/.style = {->,> = latex'}}
		\Tree
		[.{$premise$: All$^{\upred}$ black$^{\downred}$ mammals$^{\downred}$ \\ saw$^{\upred}$ exactly$^{\red{=}}$ 5$^{\red{=}}$ stallions$^{\red{=}}$ who$^{\red{=}}$ danced$^{\red{=}}$} 
		[
		.\node(inf1){\blue{Some}$^{\upred}$ black$^{\upred}$ mammal$^{\upred}$ \\ saw$^{\upred}$ exactly$^{\red{=}}$ 5$^{\red{=}}$ stallions$^{\red{=}}$ \\ who$^{\red{=}}$ danced$^{\red{=}}$};  
		[.{Some$^{\upred}$ \blue{mammal}$^{\upred}$ \\ saw$^{\upred}$ exactly$^{\red{=}}$ 5$^{\red{=}}$ stallions$^{\red{=}}$ \\ who$^{\red{=}}$ danced$^{\red{=}}$} ] 
%		[.{...} ] 
		] 
		[.\node(inf2){All$^{\upred}$ black$^{\downred}$ \blue{dogs}$^{\downred}$ \\ saw$^{\upred}$ exactly$^{\red{=}}$ 5$^{\red{=}}$ stallions$^{\red{=}}$ \\ who$^{\red{=}}$ danced$^{\red{=}}$}; 
		[.{All$^{\upred}$ black$^{\downred}$ \blue{doodles}$^{\downred}$ \\ saw$^{\upred}$ exactly$^{\red{=}}$ 5$^{\red{=}}$ stallions$^{\red{=}}$ \\ who$^{\red{=}}$ danced$^{\red{=}}$} ] 
%		[.{...} ] 
		]
%		[.\node(inf3){All$^{\upred}$ black$^{\downred}$ \blue{poodles}$^{\downred}$ \\ saw$^{\upred}$ exactly$^{\red{=}}$ 5$^{\red{=}}$ stallions$^{\red{=}}$ \\ who$^{\red{=}}$ danced$^{\red{=}}$}; 
%		[.{\blue{A}$^{\upred}$ black$^{\upred}$ poodle$^{\upred}$ \\ saw$^{\upred}$ exactly$^{\red{=}}$ 5$^{\red{=}}$ stallions$^{\red{=}}$ who$^{\red{=}}$ danced$^{\red{=}}$} 
%		[.{A$^{\upred}$ \blue{brown or black}$^{\upred}$ poodle$^{\upred}$ \\ saw$^{\upred}$ exactly$^{\red{=}}$ 5$^{\red{=}}$ stallions$^{\red{=}}$ who$^{\red{=}}$ danced$^{\red{=}}$} ]
%		]
%		]
		]
%		\node [rectangle,draw,below left=1.5cm and 0.2cm of inf1] (contra1) {\footnotesize \textit{no} black mammals \\ \footnotesize saw exactly 5 stallions who danced};
%		\node [rectangle,draw,below left=1.5cm and 0.2cm of inf2] (contra2) {\footnotesize ...};
%		\node [rectangle,draw,below left=1.5cm and 0.2cm of inf3] (contra3) {\footnotesize ...};
%		\draw[edge, shorten >=2pt, shorten <=2pt] (inf1) to (contra1);
%		\draw[edge, shorten >=2pt, shorten <=2pt] (contra1) -- node[above, rotate=30] {\tiny contradiction} (inf1) ;
%		\draw[edge, shorten >=2pt, shorten <=2pt] (inf2) to (contra2);
%		\draw[edge, shorten >=2pt, shorten <=2pt] (contra2) -- node[above, rotate=45] {\tiny contradiction} (inf2) ;
%		\draw[edge, shorten >=2pt, shorten <=2pt] (inf3) to (contra3);
%		\draw[edge, shorten >=2pt, shorten <=2pt] (contra3) -- node[above, rotate=40] {\tiny contradiction} (inf3) ;
		%\draw[semithick,->] (t)..controls +(south west:5) and +(south:5)..(wh);
		\end{tikzpicture}
		}%
		\caption{Generating \texttt{ENTAILMENT} for monotonicity fragments starting from the $premise$ (top). Each node in the tree shows an entailment generated by one \emph{substitution} (in blue). Substitutions are based on a hand-coded knowledge base with information such as: 
		\textit{all} $\leq$ \textit{some/a}, 
		\textit{poodle} $\leq$ \textit{dog} $\leq$ \textit{mammal}, 
		and \textit{black mammal} $\leq$ \textit{mammal}. \texttt{CONTRADICTION} examples are generated for each inference using simple rules such as ``replace \textit{some/many/every} in subjects by \textit{no}''. \texttt{NEUTRAL}s are generated in a reverse manner as the entailments. 
			\label{fig:gen:mono} }
		\end{figure}
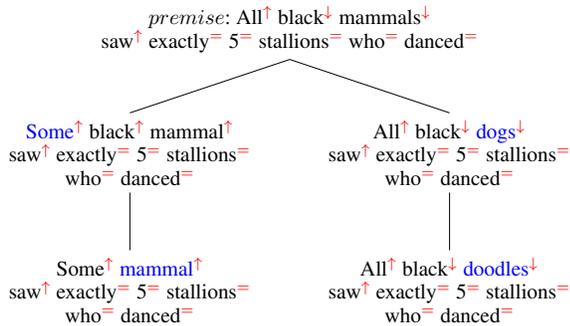

\paragraph{The Monotonicity Fragments} The second set of fragments cover monotonicity reasoning, as first discussed in the introduction. This fragment can be described using a regular grammar with polarity facts according to the monotonicity calculus, such as the following: \textit{every} is \textit{downward} monotone/entailing in its first argument but \textit{upward} monotone/entailing in the second, denoted by the $^{\textdownarrow}$ and $^{\textuparrow}$ arrows in the example sentence \textit{every$^{\textuparrow}$ small$^{\textdownarrow}$ dog$^{\textdownarrow}$ ran$^{\textuparrow}$}. We have manually encoded monotonicity information for 14 types of quantifiers (\textit{every}, \textit{some}, \textit{no}, \textit{most}, \textit{at least 5}, \textit{at most 4}, etc.) and negators (\textit{not}, \textit{without}) and generated sentences using a simple regular grammar and a small lexicon of about 100 words. We then use the system described by \citet{hu2018polarity}\footnote{\url{https://github.com/huhailinguist/ccg2mono}} to automatically assign arrows to every token (see Figure \ref{fig:gen:mono}, note that = means 
that the inference is \emph{neither} monotonically up or down in general). Because we manually encoded the monotonicity information of each token in the lexicon and built sentences via a controlled set of grammar rules, the resulting arrows assigned by \citet{hu2018polarity} can be proved to be correct. 

Once we have the sentences with arrows, we use the algorithm of \citet{HuChenMoss} to generate \textit{pairs} of sentences with \texttt{ENTAIL}, \texttt{NEUTRAL} or \texttt{CONTRADICTORY} relations, as exemplified in Figure \ref{fig:gen:mono}. Specifically, we first define a \textit{knowledge base} that stores the relations of the lexical items in our lexicon, e.g., 
\textit{poodle} $\leq$ \textit{dog} $\leq$ \textit{mammal} $\leq$ \textit{animal};
also, \textit{waltz} $\leq$ \textit{dance} $\leq$ \textit{move};
and \textit{every} $\leq$ \textit{most} $\leq$ \textit{some} $=$ \textit{a}.
For nouns,  $\leq$ can be understood as the subset-superset relation.
For higher-type objects like the determiners above, see~\citet{IcardMoss2013}
for discussion. Then to generate entailments, we perform \emph{substitution} (shown in Figure~\ref{fig:gen:mono} in blue). That is, we substitute upward entailing tokens or constituents with something ``greater than or equal to'' ($\geq$) them, or downward entailing ones with something ``less than or equal to'' them. To generate neutrals, substitution goes the reverse way. For example, \textit{all$^{\textuparrow}$ dogs$^{\textdownarrow}$ danced$^{\textuparrow}$} \texttt{ENTAIL} \textit{all poodles danced}, while \textit{all$^{\textuparrow}$ dogs$^{\textdownarrow}$ danced$^{\textuparrow}$} \textit{NEUTRAL} \textit{all mammals danced}. This is due to the facts which we have seen:
\textit{poodle} $\leq$ \textit{dog} $\leq$ \textit{mammal}. Simple rules such as ``replace \textit{some/many/every} in subjects by \textit{no}'' or ``negate the main verb'' are applied to generate contradictions. 

Using this basic machinery, we generated two separate challenge datasets, one with limited complexity (e.g., each example is limited to 1 relative clause and uses an inventory of 5 quantifiers), which we refer to throughout as \texttt{monotonicity (simple)}, and one with more overall quantifiers and substitutions, or \texttt{monotonicity (hard)} (up to 3 relative clauses and a larger inventory of 14 unique quantifiers). Both are defined over the same set of lexical items (see Figure~\ref{fig:frags}).   

 %We only use a very small lexicon as a starting point to see whether BERT and other neural models can reliably learn the inferences based on monotonicity calculus, a central part of natural logic \cite{vanBenthemHistory08}.   % As earlier, the testing vocabulary was disjoint from the training vocabulary.

		%% to add after the fact (AAAI doesn't allow supplementary data) 
		%\red{See details in text and supplementary file.} 

%\begin{align*}
%    S \to 
%\end{align*}

%The ``start symbol'' of this grammar is
%$\Sent^{\upred}$.   The grammar does not quite
%generate English, due to issues with singular/plural agreement, phonology, and related matters.   But it still generates polarized strings which are appropriate for us.
%For example, the following are generated:
%\[
%\mbox{\emph{all}}^{\upred}\
%\mbox{\emph{dog}}^{\downred} \
%\mbox{\emph{see}}^{\upred}\
%\mbox{\emph{some}}^{\upred}\
%\mbox{\emph{chase}}^{\downred}\
%\mbox{\emph{no}}^{\downred}\
%\mbox{\emph{cat}}^{\downred}
%\]

\section{Experimental Setup and Methodology}
\label{sec:exp}

To address the questions in Figure~\ref{fig:first_example}, we experiment with two task-specific NLI models from the literature, the \textbf{ESIM} model of \citet{chen2017enhanced} and the decomposable-attention (\textbf{Decomp-Attn}) model of \citet{parikh2016decomposable} as implemented in the AllenNLP toolkit \cite{gardner2018allennlp}, and the pre-trained \textbf{BERT} architecture of \citet{devlin2018bert}.\footnote{We use the \textbf{BERT-base} uncased model in all experiments, as implemented in HuggingFace: \url{https://github.com/huggingface/pytorch-pretrained-BERT}.}

When evaluating whether fragments can be learned from scratch (Question 1), we simply train models on these fragments directly using standard training protocols. To evaluate pre-trained NLI models on individual fragments (Question 2), we train BERT models on combinations of the SNLI and MNLI datasets from GLUE \cite{wang2018glue}, and use pre-trained ESIM and Decomp-Attn models trained on MNLI following \citet{liu2019inoculation}.

\begin{figure}[t!]
\centering
\begin{tabular}{c}
\includegraphics[scale=0.45]{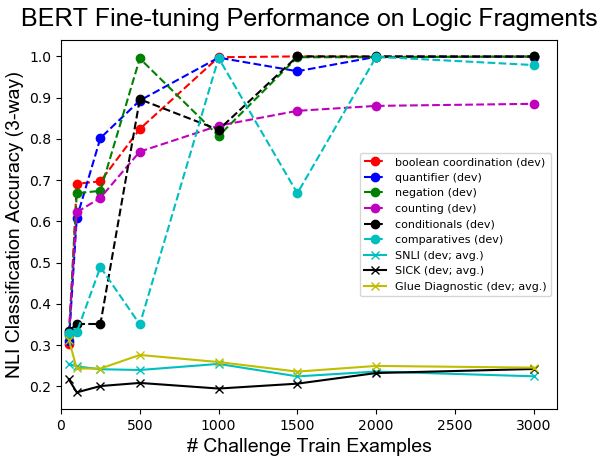} \\
\includegraphics[scale=0.45]{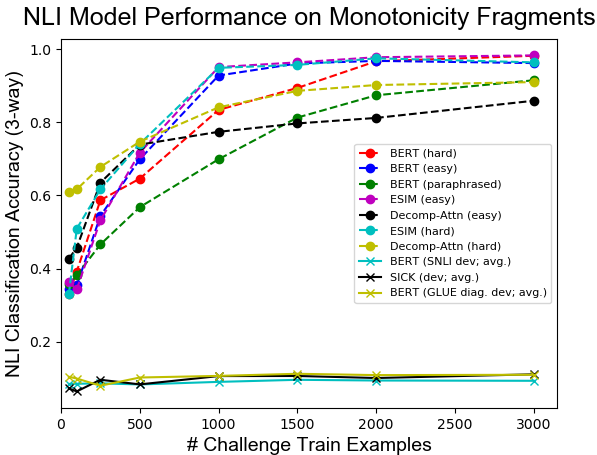}
\end{tabular}
\caption{Dev. results on training NLI models from scratch on the different fragments and architectures.}
%\caption}
\label{fig:scratch}
\end{figure}

To evaluate whether a pre-trained NLI model can be re-trained to improve on a fragment (Question 3), we employ the recent \emph{inoculation by fine-tuning} method~\cite{liu2019inoculation}. The idea is to re-fine-tune (i.e., continue training) the models above using $k$ pieces of fragment training data, where $k$ ranges from 50 to 3,000 (i.e., a very small subset of the fragment dataset to the full training set; see horizontal axes in Figures~\ref{fig:scratch},~\ref{fig:inoculation-highlights}, and~\ref{fig:inoculation-additional}). The intuition is that by doing this, we see the extent to which this additional data makes the model more robust to handle each fragment, or stresses it, resulting in performance loss on its original benchmark. In contrast to re-training models from scratch with the original data augmented with our fragment data, fine-tuning on only the new data is substantially faster, requiring in many cases only a few minutes. This is consistent with our requirement discussed previously that training existing models to be robust on new fragments should be \emph{quick}, given the multitude of fragments that we expect to encounter over time. For example, in coming up with new linguistic fragments, we might find newer fragments that are not represented in the model; it would be prohibitive to re-train the model each time entirely from scratch with its original data (e.g., the 900k+ examples in SNLI+MNLI) augmented with the new fragment. 

\newcommand{\scorefrag}{S_\text{frag}\big(M^{a,k}_j\big)}
\newcommand{\scoreorig}{S_\text{orig}\big(M^{a,k}_j\big)}

%(e.g., in coming up with a new linguistic fragment, such as the ones discussed already we might find new fragments that are not represented in the model. Given ).
Our approach to inoculation, which we call \emph{lossless inoculation}, differs from \citet{liu2019inoculation} in \emph{explicitly} optimizing the aggregate score of each model on both its original and new task. More formally, let $k$ denote the number of examples of fragment data used for fine-tuning. Ideally, we would like to be able to fine-tune each pre-trained NLI model architecture $a$ (e.g., BERT) to learn a new fragment perfectly with a minimal $k$, while---importantly---not losing performance on the original task that the model was trained for (e.g., SNLI or MNLI). Given that fine-tuning is sensitive to hyper-parameters,\footnote{We found all models to be sensitive to learning rate, and performed comprehensive hyper-parameters searches to consider different learning rates, \# iterations and (for BERT) random seeds.} we use the following methodology: For each $k$ we fine-tune $J$ variations of a model architecture, denoted $M^{a,k}_j$ for $j \in \{1, \ldots, J\}$, each characterized by a different set of hyper-parameters. We then identify a model $M^{a,k}_*$ with the best \emph{aggregated} performance based on its score $\scorefrag$ on the fragment dataset and $\scoreorig$ on the original dataset. For simplicity, we use the average of these two scores as the aggregated score.\footnote{Other ways of aggregating the two scores can be substituted. E.g., one could maximize $\scorefrag$ while requiring that $\scoreorig$ is not much worse relative to when the model's hyperparameters are optimized directly for the original dataset.} Thus, we have:
%
%% Option A:
% \begin{gather*}
%     M^{a,k}_* = M^{a,k}_{j^*} \\
%     j^* = \argmax_{j}\, \textsc{avg}\bigg(\scorefrag, \scoreorig\bigg)
% \end{gather*}
%
%% Option B:
\begin{align*}
    M^{a,k}_* & = \!\!\!\!\!\!\! \argmax_{M \in \{M^{a,k}_1, \ldots, M^{a,k}_J\}}\!\!\! \textsc{avg}\bigg(S_\text{frag}(M), S_\text{orig}(M)\bigg)
\end{align*}
By keeping the hyper-parameter space consistent among all fragments, the point is to observe how certain fragments behave relative to one another.

%The interesting part is to see if differences exists between different fragments; some might require a small $k$ to master, but might require trading off 

%In the best case, for each model $M$ and model architecture $a$ (e.g., BERT), and for each $k$, we want to be able to maximize our model's score on the fragment $S-\texttt{frag}_{M}^{(k)}$ and on the original task that the model was trained on $S_{\texttt{orig}}_{M}^{(k)}$,  
%we want to be able to maximize our model's score 
%model's score on each fragments using the smallest $k$ amount of examples without hurting the performance on the original task that the model was trained for. Given the sensitivity of model fine-tuning performance to hyper-parameters, for each $k$ we train several models $M$ characterized by different hyper-parameter settings, and find the model $M_{k}^{*}$ that, as given by the following: 

%Consistent with our point in Section~\ref{sec:intro}, such an approach, in contrast to training pre-training 

%We assume that the model's behavior after inoculation coming from a given fragment reflects its ability to handle that fragment.

%This is precisely what we aim to probe in this paper.    
%\begin{align*}
%\argmax 
%\end{align*}

\paragraph{Additional Baselines} To ensure that the challenge datasets that are generated from our fragments are not trivially solvable and subject to annotation artifacts, we implemented variants of the \textbf{Hypothesis-Only} baselines from \citet{poliak2018hypothesis}, as shown at the top of  Table~\ref{table:frag2_results}. This involves training a single-layered \textbf{biLSTM} encoder for the hypothesis side of the input, which generates a representation for the input using max-pooling over the hidden states, as originally done in \citet{conneau2017supervised}. We used the same model to train a \textbf{Premise-Only} model that instead uses the premise text, as well as an encoder that looks at both the premise and hypothesis (\textbf{Premise+Hyp.}) separated by an artificial token (for more baselines, see \citet{salvatore2019using}). 

%In all cases, such models did poorly on our fragments datasets, and in no case achieved a result close to perfect accuracy (in three of the six logic fragments, the accuracy numbers came close to random chance at $33\%$), which one would expect if the datasets were trivially solvable. For more baselines on earlier versions of these logic fragments, see \citet{salvatore2019using}). 

%, given in different amounts or \emph{doses}. The intuition is that by doing this, we see the extent to which this additional data fragment (i.e., the \emph{inoculation} or vaccine) fixes and/or stresses (e.g., causes it to lose performance on its original task) the model. We assume that the model's behavior after inoculation coming from a given fragment reflects its ability to handle that fragment.
%This is precisely what we aim to probe
%in this paper.

%. In their original work, this was used to evaluate the difficult of challenge datasets. Our idea is to use this technique for evaluating more specifically the difficulty of target linguistic fragments and phenomena.  

%For evaluating pre-trained models on our new challenge sets, we train BERT using 

%To find differences between the above fragments and 
%We fine-tune BERT using the publicly-available Huggingface\footnote{\url{https://github.com/huggingface/pytorch-pretrained-BERT}} PyTorch implementation

\begin{table*}[t]
    \centering
    \footnotesize
    \scalebox{0.9}{
    \begin{tabular}{|  l c c c c |}
    \hline\hline
    \textbf{Model$_{\texttt{train\_data}}$} & \textbf{SNLI Test} & \textbf{Logic Fragments (Avg. of 6)} & \textbf{Mono. Fragments (Avg. over 2)} & \textbf{Breaking NLI} \\ \hline\hline
    \multicolumn{5}{|c|}{\textbf{Random/Trained Baselines}}  \\ \cdashline{1-5}
    \textbf{Majority Baseline} & \multicolumn{1}{c}{34.2} & 34.6 & 34.0 & --  \\
    \textbf{Hypothesis-Only biLSTM} & \multicolumn{1}{c}{69.0} & 49.3 & 56.7 & --  \\
    \textbf{Premise-Only biLSTM} & \multicolumn{1}{c}{--} &  44.3 & 57.4 &  -- \\
    \textbf{Premise+Hyp. biLSTM} & \multicolumn{1}{c}{--} & 52.0 & 59.1 &  -- \\
    \multicolumn{5}{|c|}{\textbf{Pre-Trained NLI Models}} \\ \cdashline{1-5}
    \textbf{BERT}$_{\texttt{SNLI+MNLI}}$ & \multicolumn{1}{c}{91.0} & 47.3 & 62.8  & 95.8 \\ 
    \textbf{BERT}$_{\texttt{SNLI}}$ & \multicolumn{1}{c}{90.7} & 46.1 & 56.8  & 94.3 \\ 
    \textbf{Decomp-Attn}$_{\texttt{SNLI}}$ & \multicolumn{1}{c}{86.4} & 42.1 & 48.4  & 49.9 \\ %\hline % \hline
    \textbf{ESIM}$_{\texttt{SNLI}}$ & \multicolumn{1}{c}{88.5} &  44.3 & 62.8 & 68.7 \\ 
    & \textbf{MNLI Dev (Avg.)} & \multicolumn{3}{c|}{\textbf{Re-Trained Models with Fragments (\texttt{frag})}} \\ \cdashline{1-5}
    \textbf{BERT}$_{\texttt{SNLI+MNLI+frag}}$ & \multicolumn{1}{c}{83.7}$(\downarrow 1.3)$ & 98.0 & 97.8  & - \\
    \textbf{ESIM}$_{\texttt{MNLI+frag}}$ & \multicolumn{1}{c}{72.0}$(\downarrow 5.9)$ & 86.4 & 96.5  & - \\
    \textbf{Decomp-Attn}$_{\texttt{MNLI+frag}}$ & \multicolumn{1}{c}{66.1}$(\downarrow 6.7)$ & 71.7 & 93.5  & - \\ \hline
    %\multicolumn{5}{|c|}{\textbf{Re-Trained Models}} \\ \cdashline{1-5}
    %\multicolumn{5}{|c|}{} \\ \hline
    %\multicolumn{4}{|c|}{\textbf{BERT Re-trained with Orig.+Fragment Data}} \\ 
    %BERT$_{\texttt{snli+mnli+chall.}}$ & 91.2 (\textcolor{green}{+0.2}) &  & \\ \hline 
    \end{tabular}}
    %}
    \caption{Baseline models and model performance (accuracy \%) on NLI benchmarks and challenge test sets (before and after re-training), including the \textbf{Breaking NLI} challenge set from \citet{glockner2018breaking}. The arrows $\downarrow$ in the last section show the average drop in accuracy on MNLI benchmark after re-training with the fragments. }
    \label{table:frag2_results}
\end{table*}

\section{Results and Findings}
\label{sec:findings}
We discuss the different questions posed in Figure~\ref{fig:first_example}. 

%In this section, we report on the results found for the particular fragments introduced above. 

\subsubsection{Answering Questions 1 and 2.}

% All 4 models can master Monotonicity Reasoning while retaining accuracy on their original benchmarks. However, non-BERT models lose substantial accuracy on their original benchmark when trying to learn Comparatives.

%\todo{adjust discussion in text accordingly}

%\ashish{I split the big figure into two pieces, and added a self-explanatory caption.}

Table~\ref{table:frag2_results} shows the performance of baseline models and pre-trained NLI models on our different fragments. In all cases, the baseline models did poorly on our datasets, showing the inherent difficulty of our challenge sets. In the second case, we see clearly that state-of-the-art models do not perform well on our fragments, consistent with findings on other challenge datasets. One result to note is the high accuracy of BERT-based pre-trained models on the \textbf{Breaking NLI} challenge set of \citet{glockner2018breaking}, which previously proved to be a difficult benchmark for NLI models. This result, we believe, highlights the need for more challenging NLI benchmarks, such as our new datasets. 

%In all cases, such models did poorly on our fragments datasets, and in no case achieved a result close to perfect accuracy (in three of the six logic fragments, the accuracy numbers came close to random chance at $33\%$), which one would expect if the datasets were trivially solvable. For more baselines on earlier versions of these logic fragments, see \citet{salvatore2019using}). 

% \begin{figure*}
% \centering
% \begin{tabular}{c c c}
%      \includegraphics[scale=.3]{bool_inoc.png} &\includegraphics[scale=.3]{negation_inoc.png} & \includegraphics[scale=.3]{quantifier_inoc.png}  \\
%      \includegraphics[scale=.3]{counting_inoc.png} &
%      \includegraphics[scale=.3]{conditional_inoc.png} & 
%      \includegraphics[scale=.3]{comparatives_inoc.png} \\
%      \includegraphics[scale=.3]{mono_easy_inoc.png} & \includegraphics[scale=.3]{mono_hard_inoc.png} 
%      & \includegraphics[scale=.6]{legend.png} \\
%      %\includegraphics[scale=.37]{fragment3_plot.png} &
%      %\includegraphics[scale=.37]{mixed_fragment_plot.png} &
%      %\includegraphics[scale=.37]{all_fragments_plot.png}
% \end{tabular}
% \caption{Inoculation results for the different fragments, where each point represents the best model $M_{k}^{*}$.}
% \label{fig:inoculation}
% \end{figure*}
\begin{figure*}[t]
\centering
    \includegraphics[width=0.35\textwidth]{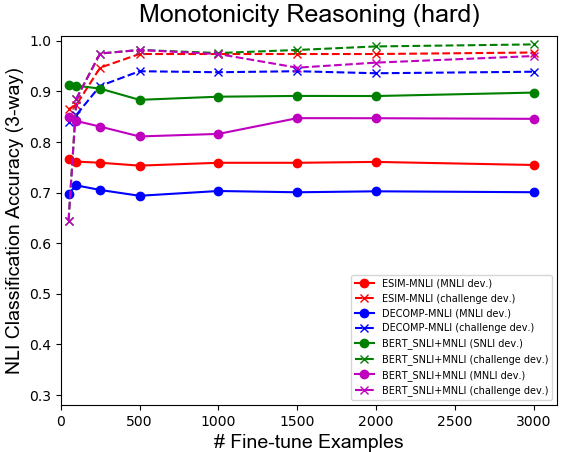}
    \hspace{5ex}
    \includegraphics[width=0.35\textwidth]{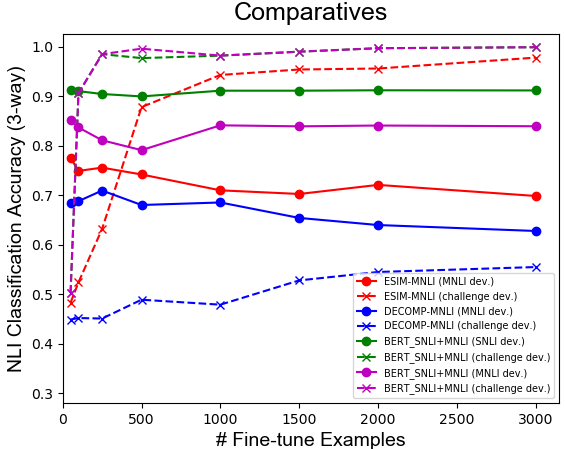}
\caption{Inoculation results for two illustrative semantic fragments, Monotonicity Reasoning (left) and Comparatives (right), for 4 NLI models shown in different colors. Horizontal axis: number of fine-tuning challenge set examples used. Each point represents the model $M_{k}^{*}$ trained using hyperparameters that maximize the accuracy averaged across the model's original benchmark dataset (solid line) and challenge dataset (dashed line).}
\label{fig:inoculation-highlights}
\end{figure*}

\begin{figure*}[t]
\centering
     \includegraphics[scale=.32]{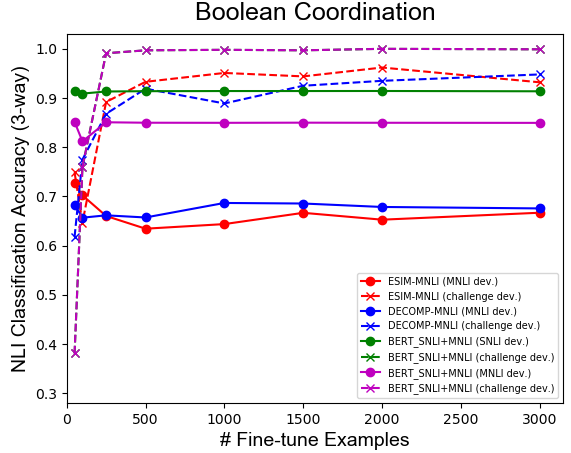}
     \hspace{2ex}
     \includegraphics[scale=.32]{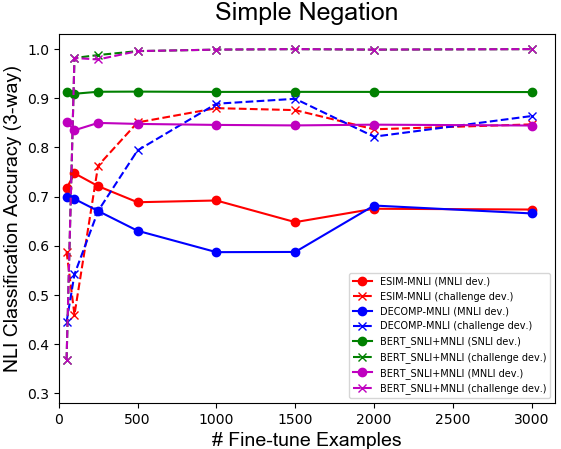}
     \hspace{2ex}
     \includegraphics[scale=.32]{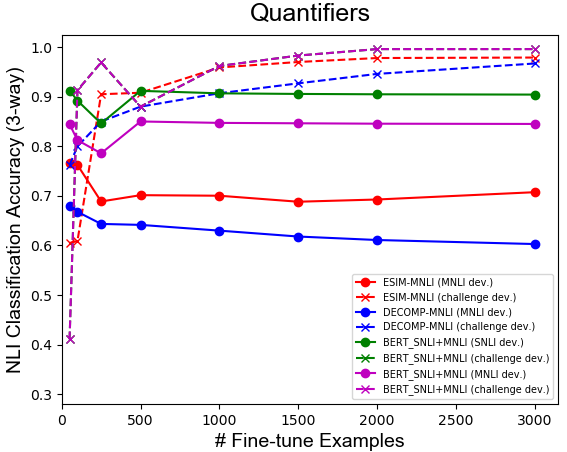}
     \\
     \includegraphics[scale=.32]{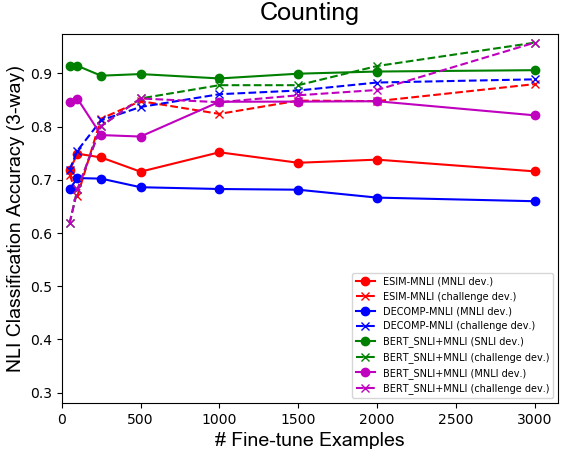}
     \hspace{2ex}
     \includegraphics[scale=.32]{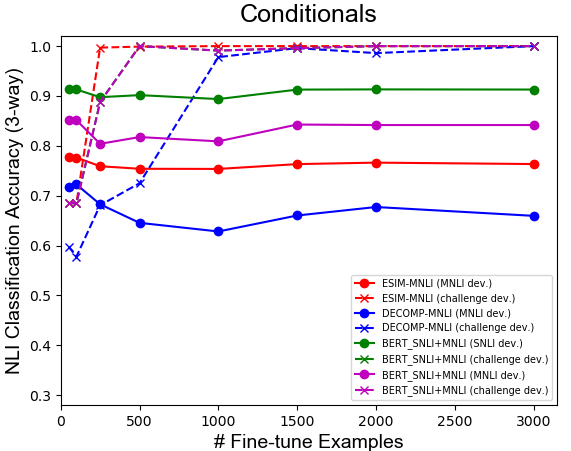}
     \hspace{2ex}
     \includegraphics[scale=.32]{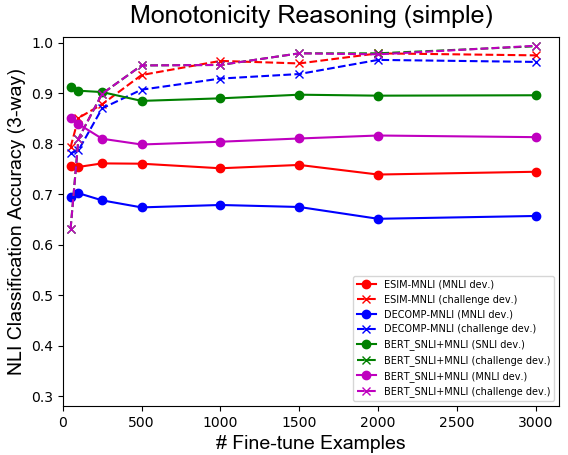}
\caption{Inoculation results for 6 semantic fragments not included in Figure~\ref{fig:inoculation-highlights}, using the same setup.}
\label{fig:inoculation-additional}
\end{figure*}

Figure~\ref{fig:scratch} shows the results of training NLI models from scratch (i.e., without NLI pre-training on other benchmarks) on the different fragments. In nearly all cases, it is possible to train a model to master a fragment (with \texttt{counting} being the hardest fragment to learn). In other studies on learning fragments \cite{geiger2018stress,salvatore2019using}, this is the main result reported, however, we also show that the resulting models perform below random chance on benchmark tasks, meaning that these models are not by themselves very useful for general NLI. This even holds for results on the GLUE diagnostic test \cite{wang2018glue}, which was hand-created and designed to model many of the logical phenomena captured in our fragments. 

We note that in the monotonicity examples, we included results on a development set (in dashed green) that was built by systematically paraphrasing all the nouns and verbs in the fragment to be disjoint from training. Even in this case, when lexical variation is introduced, the BERT model is robust (see \citet{rozen2019analyzing} for a more systematic study of this type of generalization using BERT for NLI in different settings). 

%Results in Table~\ref{table:frag2_results}. 

\subsubsection{Answering Question 3.}

Figures~\ref{fig:inoculation-highlights} and~\ref{fig:inoculation-additional} show the results of the re-training  study. They compare the performance of a retrained model on the challenge tasks (dashed lines) as well as on its original benchmark tasks (solid lines)\footnote{For \textbf{MNLI}, we report results on the mismatched dev.~set.}. We discuss here results from the two illustrative fragments depicted in Figure~\ref{fig:inoculation-highlights}. All 4 models can master Monotonicity Reasoning while retaining accuracy on their original benchmarks. However, non-BERT models lose substantial accuracy on their original benchmark when trying to learn \texttt{comparatives}, suggesting that comparatives are generally harder for models to learn. In Figure~\ref{fig:inoculation-additional}, we show the results for all other fragments, which show varied, though largely stable, trends depending on the particular linguistic phenomena. 

At the bottom of Table~\ref{table:frag2_results}, we show the resulting  accuracies on the challenge sets and MNLI bechmark for each model after re-training (using the optimal model $M^{a,k}_*$, as described previously). In the case of \textbf{BERT}$_{\texttt{SNLI+MNLI+frag}}$, we see that despite performing poorly on these new challenge dataset before re-training, it can learn to master these fragments  with minimal losses to performance on its original task (i.e., it only loses on average about 1.3\% accuracy of the original MNLI dev set). In other words, it is possible teach BERT (given its inherent capacity) a new fragment quickly through re-training without affecting its original performance, assuming however that time is spent on carefully finding the optimal model.\footnote{We note that models without optimal aggregate performance are often prone to catastrophic forgetting.} For the other models, there is more of a trade-off; \textbf{Decomp-Attn} on average never quite masters the logic fragments (but does master the \texttt{Monotonicity Fragments}), and incurs an average 6.7\% loss on MNLI after re-training. In the case of comparatives, the inability of the model to master this fragment likely reveals a certain architectural limitation of the model given that it is not sensitive to word-order. Given such losses, perhaps in such cases a more sophisticated re-training scheme is needed in order to optimally learn particular fragments.

%\ashish{update the following to focus on and contrast the two illustrative fragments. Highlight the stability for Monotonicity Reasoning, and the lack of it (plus the difficulty faced by Decomp-Attn in mastering the fragment) for Comparatives.}
%In all cases, the results on the original tasks deteriorate as the model improves on the fragment, however there are marked differences between the levels of deterioration. In the case of quantifiers, an \textbf{SNLI+MNLI} trained model, which is trained on nearly 1 million examples can lose around 30\% of its accuracy on SNLI after only 500 new examples. (In other fragments, the trend is similar though less dramatic.)
%In contrast, all models remain relatively stable for the counting and monotonicity reasoning fragments. 

%. For each fragment, we see very different behavior. In the case of quantifiers, we see that there is a major trade-off for all NLI models between mastering the 

%Results shown in plots Figure~\ref{fig:inoculation}, as partly in Table~\ref{table:frag2_results} (last results are still missing, will be filled in tomorrow). 

\section{Discussion and Conclusion}

We explored the use of \emph{semantic fragments}---systematically controlled subsets of language---to probe into NLI models and benchmarks. Our investigation considered 8 particular fragments and new challenge datasets that center around basic logic and monotonicity reasoning. In answering the questions first introduced in Figure~\ref{fig:first_example}, we found that while existing NLI architectures are able to learn these fragments from scratch, the resulting models are of limited interest. Further, pre-trained models perform poorly on these new datasets (even relative to other available challenge benchmarks), revealing the weaknesses of these models. Interestingly, however, we show that many models can be quickly re-tuned (e.g., often in a matter of minutes) to master these different fragments using a novel variant of the \emph{inoculation through fine-tuning} strategy~\cite{liu2019inoculation} that we introduce called \emph{lossless inoculation}. 

Our results suggest the following methodology for improving models: Given a particular linguistic hole in an NLI model, one can plug this hole by simply generating synthetic data and using it to re-train a model. This methodology comes with some caveats, however: Depending on the model and particular linguistic phenomena, there may be some trade-offs with the model's original performance, which should first be looked at empirically and compared against other linguistic phenomena. Our work is one small step in trying to gather an inventory of NLI phenomena and look rigorously at model performance, which follows earlier work on NLI (see \citet{zaenen2005local}).

\paragraph{Can we find more difficult fragments?}
Despite differences across various fragments, we largely found NLI models to be robust when tackling new linguistic phenomena and easy to quickly re-purpose (especially with BERT). This generally positive result begs the question: Are there more challenging fragments and linguistic phenomena that we should be studying?

The ubiquity of logical and monotonicity reasoning provides a justification for our particular fragments, and we take it as a positive sign that models are able to solve these tasks. As we emphasize throughout, however, our general approach is amenable to any linguistic phenomena, and future work may focus on developing more complicated fragments that capture a wider range of linguistic phenomena and inference. This could include, for example, efforts to extend to fragments in a way that moves beyond elementary logic to systematically target the types of commonsense reasoning known to be common in existing NLI tasks \cite{lobue2011types}. We believe that semantic fragments are a promising way to introspect model performance generally, and can also be used to forge interdisciplinary collaboration between neural NLP research and traditional linguistics. 

%encourage the type of interdisciplinary collaboration between NLP and
%neighboring fields in language science that has diminished in recent
%years.

Benchmark NLI annotations and judgements are often imperfect and error-prone (cf.~\citet{kalouli2017correcting}, \citet{pavlick2019inherent}), partly due to the loose way in which the task is traditionally defined \cite{dagan2005pascal}. For models trained on benchmarks such as SNLI, understanding model performance not only requires probing how each target model works, but also probing the particular flavor of NLI that is captured in each  benchmark. We believe that our variant of inoculation and overall framework can also be used to more systematically look at these issues, as well as help identify annotation errors and artifacts.

\paragraph{What are Models Actually Learning?} 
One open question concerns the extent to which models trained on narrow fragments can generalize beyond them. Newer \emph{analysis methods} that attempt to correlate neural activation patterns and target symbolic patterns \cite{chrupala2019correlating} might help determine the extent to which models are truly generalizing, and provide insights into alternative ways of training more robust and generalizable models.

A key feature of our \emph{lossless} inoculation strategy, which differs from the original proposal of \citet{liu2019inoculation}, is that each time we teach the model something new, we explicitly take into account how much loss this same model has on its original task, and balance the two scores accordingly. The fact that models such as BERT can effectively learn new tasks with minimal loss on their original tasks gives some indication that, even if the models are not generalizing too far beyond the provided challenge tasks, one way to increase generalization is by continuously feeding models new challenge tasks. This type of continuous or never-ending learning scenario is one promising area
% we intend to pursue in future work
for future work that one may pursue
by looking at more robust methods for model inoculation and fine-tuning.

\section{Acknowledgements}
We thank the anonymous reviewers for their helpful feedback, as well as our colleagues, especially Peter Clark, Vered Shwartz, and Reut Tsarfaty. Part of this work is supported by grant $\#$586136 from the Simons Foundation. Hai Hu is supported by the China Scholarship Council.

\bibliography{AAAI-RichardsonK.9757.bib}
\bibliographystyle{aaai}

%% appendix taken out 
\pagebreak

\section{Appendix} 

We provide more details of how the monotonicity fragment. To generate this fragment, we first defined a grammar with a hand coded lexicon. Then we use the tool described in \citet{hu2018polarity} to obtain polarities (arrows) on each constituent of the sentences. Finally we implement the substitution algorithm on the polarized sentences described in \citet{HuChenMoss}. These steps are detailed below. 

\paragraph{Grammar and Lexicon}

All premises in the monotonicity fragment are generated using the grammar and lexicon detailed in Figure~\ref{fig:mono_gram} included in this Appendix.

\begin{figure*}
    \centering
    {\footnotesize
 
    \begin{tabular}{c}
    %{\footnotesize
    \textbf{Grammar}:\\
        \begin{tabular}{lll}
S &  $\rightarrow$  & NP${}_{animate}$ VP\\

NP &  $\rightarrow$  & NP${}_{animate}$ $\vert$ NP${}_{inanimate}$\\

NP${}_{animate}$ &  $\rightarrow$  & Det (Adj${}_{animate}$) Nbar${}_{animate}$\\

NP${}_{inanimate}$ &  $\rightarrow$  & Det (Adj${}_{inanimate}$) N${}_{inanimate}$\\

Nbar${}_{animate}$ &  $\rightarrow$  & N${}_{animate}$ $\vert$ N${}_{animate}$  SRC $\vert$ N${}_{animate}$  ORC $\vert$ N${}_{animate}$  PP\\

SRC &  $\rightarrow$  & who  VP\\

ORC &  $\rightarrow$  & who  NP${}_{animate}$  Vt\\

PP &  $\rightarrow$  & P Adj${}_{smell}$ N${}_{smell}$\\

VP &  $\rightarrow$  & VPbar $\vert$ do not VPbar $\vert$ be (not) Adj${}_{pred}$\\

VPbar &  $\rightarrow$  & V$_{i}$ $\vert$ V$_{t}$  NP\\
&& (note: SRC = subject relative clause, ORC = object relative clause)
\end{tabular} \\ \\
\textbf{Lexicon}:\\ 
{
\begin{tabular}{llp{12cm}}
Det &  $\rightarrow$  & \{every, all, each, a, many, several, 
	few, most, the, some but not all, no,
	at least n, at most n, exactly n\}\\

Adj${}_{inanimate}$ &  $\rightarrow$  & \{wooden, hardwood, metal, plastic, iron, steel\}\\

Adj${}_{animate}$ &  $\rightarrow$  & \{old, young, newborn, brown, black, brown or black\}\\

Adj${}_{pred}$   &  $\rightarrow$  &   \{happy, sad\}\\

Adj${}_{smell}$ &  $\rightarrow$  & \{strong, faint\}\\

Adj${}_{subsective}$ &  $\rightarrow$  & \{good, bad, nice\}  \# only used in building knowledge base\\

N${}_{animate}$ &  $\rightarrow$  & \{dog, cat, rabbit, animal, mammal, poodle, beagle,
    bulldog, bat, horse, stallion, badger, quadruped\}\\

N${}_{inanimate}$ &  $\rightarrow$  & \{table, wagon, chair, door, object, wheel, box,    mailbox, wheelbarrow, fence\}\\

N${}_{smell}$ &  $\rightarrow$  & \{smell, odor, scent\}\\

V$_{t}$ &  $\rightarrow$  & \{saw, stared-at, inspected, hit, touched,
       moved-towards, moved-away-from, scratched, sniffed\}\\

V$_{i}$ &  $\rightarrow$  & \{slept, ran, moved, swam, waltzed, danced\}\\

P &  $\rightarrow$  & \{with, without\}\\
\end{tabular}} \\ \\ [.4cm]

\textbf{Example Pre-order/Antonym Relations from Knowledge Base}:\\

\begin{tabular}{lp{13cm}}
\textit{adjectives} & brown $\leq$ brown or black, black $\leq$ brown or black \\
& iron $\leq$ metal, steel $\leq$ metal, steel $\leq$ iron, hardwood $\leq$ wooden\\\hline
\textit{nouns} & \\
x $\leq$ animal & for x in \{dog, cat, rabbit, mammal, poodle, beagle,
              bulldog, bat, horse, stallion, badger\} \\
x $\leq$ mammal & for x in \{dog, cat, rabbit, poodle, beagle,
  bulldog, bat, horse, stallion, badger\} \\      
x $\leq$ dog & for x in \{poodle, beagle, bulldog\}\\
x $\leq$ object & for x in N$_{inanimate}$ \\
\hline
\textit{verbs} & \\
x $\leq$ moved & for x in \{ran, swam, waltzed, danced\}\\
& stare at $\leq$ saw, hit $\leq$ touch, waltzed $\leq$ danced \\
\hline
\textit{determiners} &    every = all = each $\leq $ most $\leq $ some = a, many
$\leq$ several $\leq$ at least 3 $\leq$ at least 2 $\leq$ some = a,\\
 & no $\leq$ at most 1 $\leq$ at most 2 $\leq \cdots$\\
\hline
\textit{other rules} &
Adj N $\leq$ N, N + (SRC $\vert$ ORC) $\leq$ N, \ldots \\
\hline
\textit{antonyms} & moved-towards $\perp$ moved-away-from, x $\perp$ slept for x in \{ran, swam, waltzed, danced, moved\} \\
& at most 4 $\perp$ at least 5, exactly 4 $\perp$ exactly 5,  every $\perp$ some but not all, \ldots\\
%adj x $\leq $ x & for adj in \{wooden, hardwood, metal, plastic, iron, steel\}\\
% & and x in\{table, wagon, chair, door, object, wheel, box,    mailbox, wheelbarrow, fence\}\\
%adj${}'$ y $\leq $ y & for adj${}'$ in 
%\{old, young, newborn, brown, black, brown or black\}\\
% & and y in  \{dog, cat, rabbit, animal, mammal, poodle, beagle,
%    bulldog, bat, horse, stallion, badger, quadruped\}\\
\end{tabular} \\ \\
\end{tabular}}
    \caption{A specification of the grammar and lexicon used to generate the monotonicity fragments.}
    \label{fig:mono_gram}
\end{figure*}

\paragraph{Polarization}

We use the tool from \citet{hu2018polarity} to polarize a generated sentence. For example, if the input is \texttt{every dog slept}, their tool will output \texttt{every${}^{\uparrow}$ dog${}^{\downarrow}$ slept${}^{\uparrow}$}, represented as a CCG tree \cite{steedman}. % \cite{steedman}.

\paragraph{Substitution}

%% pseudo code 

As described in the paper and Figure~\ref{fig:gen:mono} in the paper, we follow \citet{HuChenMoss} to generate sentence pairs from an input premise. In simple terms, this algorithm uses depth-first search to expand the sets of inferences, neutrals and contradictions with respect to the premise. The main search tree (Figure~\ref{fig:gen:mono}) is based on inferences, but at each node neutrals and contradictions are also generated (see function \texttt{substitute()} below).

{\footnotesize
\begin{algorithm}%[ht]
\algnewcommand\algorithmicinput{\textbf{Input:}}
\algnewcommand\INPUT{\item[\algorithmicinput]}
\algnewcommand\algorithmicoutput{\textbf{Output:}}
\algrenewcommand\algorithmicindent{1em}
\algnewcommand{\LineComment}[1]{\State \(\triangleright\) #1}
\algnewcommand\OUTPUT{\item[\algorithmicoutput]}
\caption{Generating Inferences}
\begin{algorithmic}[1]
\INPUT  a sentence \texttt{sentence}, knowledge base $\mathcal{K}$, depth $d$, starting depth $0$. 
\OUTPUT Lists of inferences. 
\Function{Search}{$ps$,$d$}
\State $\texttt{infer, contr, neutr} \leftarrow [(0,ps)], [], []$
\While{$\textbf{len}(\texttt{infer}) > 0$}
\State $\texttt{depth},s \leftarrow \texttt{infer.pop()}$
\State $\texttt{depth } += 1$
\If{$\texttt{depth} \leq d$}
\LineComment{\textcolor{blue!60}{Generate new inferences}}
\State $e,n,c \leftarrow \textsc{substitute}(s)$ 
\State $\texttt{infer.add}((\texttt{depth},e))$
\State $\texttt{neutr.add}(n)$
\State $\texttt{contr.add}(c)$
\EndIf
\EndWhile
\State \textbf{return} \texttt{infer,contr,neutr}
\EndFunction
\Function{Substitute}{s}
\State $\texttt{infer, contr, neutr} \leftarrow [], [], []$
%% look through consituents
\LineComment{\textcolor{blue!60}{Iterate sentence constituents}}
\For{\texttt{const} in $s$}
\LineComment{\textcolor{blue!60}{If polarity is up, generalize span}}
\If{\texttt{const.polarity} is ${\upred}$}
\For{\texttt{repl} in $\mathcal{K}[\texttt{const}]\texttt{.greater}$}
\State $ns \leftarrow \textsc{Swap}(s,\texttt{const},\texttt{repl})$
\State $ps \leftarrow \textsc{Polarize}(ns)$
\State $\texttt{infer.add}(ps)$
\EndFor
\For{\texttt{repl} in $\mathcal{K}[\texttt{const}]\texttt{.less}$}
\State $ns \leftarrow \textsc{Swap}(s,\texttt{const},\texttt{repl})$
\State $ps \leftarrow \textsc{Polarize}(ns)$
\State $\texttt{neutr.add}(ps)$
\EndFor
\EndIf
\LineComment{\textcolor{blue!60}{If polarity is down, specialize span}}
\If{\texttt{const.polarity} is ${\downred}$}
\For{\texttt{repl} in $\mathcal{K}[\texttt{const}]\texttt{.less}$}
\State $ns \leftarrow \textsc{Swap}(s,\texttt{const},\texttt{repl})$
\State $ps \leftarrow \textsc{Polarize}(ns)$
\State $\texttt{infer.add}(ps)$
\EndFor
\For{\texttt{repl} in $\mathcal{K}[\texttt{const}]\texttt{.greater}$}
\State $ns \leftarrow \textsc{Swap}(s,\texttt{const},\texttt{repl})$
\State $ps \leftarrow \textsc{Polarize}(ns)$
\State $\texttt{neutr.add}(ps)$
\EndFor
\EndIf
\LineComment{\textcolor{blue!60}{Find negation replacements}}
\For{\texttt{repl} in $\mathcal{K}[\texttt{const}]\texttt{.negate}$}
\State $ns \leftarrow \textsc{Swap}(s,\texttt{const},\texttt{repl})$
\State $ps \leftarrow \textsc{Polarize}(ns)$
\State $\texttt{contr.add}(ps)$
\EndFor
\EndFor
\State \textbf{return} \texttt{infer,contr,neutr}
\EndFunction
%\State
\State $ps \leftarrow \textsc{Polarize}(\texttt{sentence})$
\State \textbf{return} \textsc{search}($ps,d$)
\end{algorithmic}
\end{algorithm}}

Algorithm 1 show the pseudocode for this procedure. Assuming an input sentence \texttt{sentence}, the model first polarizes this sentence (line 43 using the \textsc{polarize} function) using the algorithm of \citet{hu2018polarity} as described above. Three sets of sentences are created (line 2): entailments of \texttt{sentence} (\texttt{infer}), contradictions (\texttt{contr}), and neutral inferences (\texttt{neutr}). Infered sentences from \texttt{infer} are then selected (starting line 3) and new inferences are generated using the \textsc{substitute} function called on line 8.

The \textsc{Substitute} function works in the following way: for each polarized sentence $s$, it enumerates through all constituents (\texttt{const}) in $s$'s parse tree (equal to $(2\mid s\mid-1)$ consituents, since we assume binary derivation rules). For each constituent,  the polarity marking (i.e., the arrows ${\upred},{\downred}$) is used to determine how span substitution should work. For example, if the \texttt{polarity} of the constituent if up (${\upred}$, see line 18), then it will generate a span substitution (using the \textsc{Swap} function) using rules from a knowledge base $\mathcal{K}$ that generalize (or have a \texttt{greater} relation with \texttt{const}). The opposite for constitutes with $\downred$ labels.

%For more details, see \citet{HuChenMoss}.

%For more details, see \citet{HuChenMoss}.

\end{document}